\documentclass[12pt]{article}
\usepackage[utf8]{inputenc}
\usepackage[margin=1in]{geometry}
\usepackage{hyperref}
\usepackage{graphicx}
\usepackage{cite}
\usepackage{framed}
\usepackage{xcolor}
\usepackage[framemethod=tikz]{mdframed}

\newmdenv[
  backgroundcolor=gray!15,
  linecolor=black,
  linewidth=1pt,
  roundcorner=5pt
]{myshadedbox}

\colorlet{shadecolor}{gray!15}

\begin{document}

\begingroup
\renewcommand{\thefootnote}{\fnsymbol{footnote}}
\footnotetext[1]{Preprint. Work in progress}
\endgroup

\begin{quote}
\begin{center}
    {
    \vspace{1em}
    
    \rule{0.8\textwidth}{4pt}
    \vspace{1em}
    
    \Large \bf DermaSynth: Rich Synthetic Image-Text Pairs Using Open Access Dermatology Datasets}\\[1em]
    
    \rule{0.8\textwidth}{2pt}  
    \vspace{1em}

    \normalsize
    \textbf{Abdurrahim Yilmaz}\textsuperscript{1,*},
    \textbf{Furkan Yuceyalcin}\textsuperscript{2},
    \textbf{Ece Gokyayla}\textsuperscript{3},
    \textbf{Donghee Choi}\textsuperscript{1},
    \textbf{Ozan Erdem}\textsuperscript{4},
    \textbf{Ali Anil Demircali}\textsuperscript{1},
    \textbf{Rahmetullah Varol}\textsuperscript{5},
    \textbf{Ufuk Gorkem Kirabali}\textsuperscript{2},
    \textbf{Gulsum Gencoglan}\textsuperscript{6},
    \textbf{Joram M. Posma}\textsuperscript{1,*},
    \textbf{Burak Temelkuran}\textsuperscript{1,*}\\[0.5em]

    {\footnotesize
    \textsuperscript{1}Imperial College London, 
    \textsuperscript{2}Yildiz Technical University,
    \textsuperscript{3}Usak Research and Training Hospital,
    \textsuperscript{4}Istanbul Medeniyet University,
    \textsuperscript{5}Universität der Bundeswehr München,
    \textsuperscript{6}Istanbul Medicana Atakoy Hospital
    }\\[1em]

    {\footnotesize \textsuperscript{*}\{a.yilmaz23, j.posma11, b.temelkuran\}@imperial.ac.uk
}


\vspace{1em}

\vspace{1em}
\noindent
\textbf{Abstract}
\end{center}
A major barrier to developing vision large language models (LLMs) in dermatology is the lack of large image--text pairs dataset. We introduce \emph{DermaSynth}, a dataset comprising of 92,020 synthetic image--text pairs curated from 45,205 images (13,568 clinical and 35,561 dermatoscopic) for dermatology-related clinical tasks. Leveraging state-of-the-art LLMs, using Gemini 2.0, we used clinically related prompts and self-instruct method to generate diverse and rich synthetic texts. Metadata of the datasets were incorporated into the input prompts by targeting to reduce potential hallucinations. The resulting dataset builds upon open access dermatological image repositories (DERM12345, BCN20000, PAD-UFES-20, SCIN, and HIBA) that have permissive CC-BY-4.0 licenses. We also fine-tuned a preliminary Llama-3.2-11B-Vision-Instruct model, DermatoLlama 1.0, on 5,000 samples. We anticipate this dataset to support and accelerate AI research in dermatology. Data and code underlying this work are accessible at \url{https://github.com/abdurrahimyilmaz/DermaSynth}.
\end{quote}

\section*{1. Introduction}
Instruction-following method has emerged as a powerful technique for leveraging large language models (LLMs) as teacher models to generate synthetic data, which can then be used to optimise more specialised models. By prompting advanced models such as GPT-4 and GPT-4 vision, researchers can efficiently create high-quality instruction-following demonstrations that significantly boost model performance for both text \cite{peng2023instruction} and vision cases \cite{liu2024visual, liu2024improved}. However, while a few specialised models have started to emerge for medical tasks \cite{wu2024pmc, han2023medalpaca} and vision medical tasks \cite{shu2023visual, li2024llava}, vision LLMs in medical domain are still new and faces numerous constraints—including privacy considerations and limited public datasets. Due to these constraints, medical vision LLMs do not yet have large scale datasets that are available in general domain \cite{alpaca}. A promising approach to mitigating these data limitations is self-instruct~\cite{selfinstruct}, a method that enhances a pretrained language model’s instruction-following capabilities by automatically generating new instructions from its own outputs. 

Among medical fields that rely on image analysis, dermatology poses a unique data challenge. Unlike radiology or histopathology—where clinical reports or slide annotations often accompany images—dermatological cases frequently lack formal textual documentation beyond brief physician annotations. This scarcity of paired image–text data complicate the training/fine-tuning and evaluation of vision LLMs, specifically targeting dermatological tasks, such as lesion classification or descriptive question answering. This gap hinders the broader research community, which is eager to explore how advanced language-and-vision models can transform clinical decision-making and improve patient outcomes.

To address these needs, we introduce DermaSynth, a collection of 92,020 synthetic image–text pairs focusing on dermatology. Our dataset leverages instruction-following methods—combining state-of-the-art models with expert selected prompts—to produce rich textual annotations for a variety of dermatology images. By bridging the gap between visual data and clinically relevant text, DermaSynth enables researchers to develop, test, and refine vision-based instruction-following models tailored to dermatology. This advancement enhances both model interpretability and diagnostic reliability in an underrepresented yet critical medical domain.

\begin{figure}[!ht]
    \centering
    \includegraphics[width=1\textwidth]{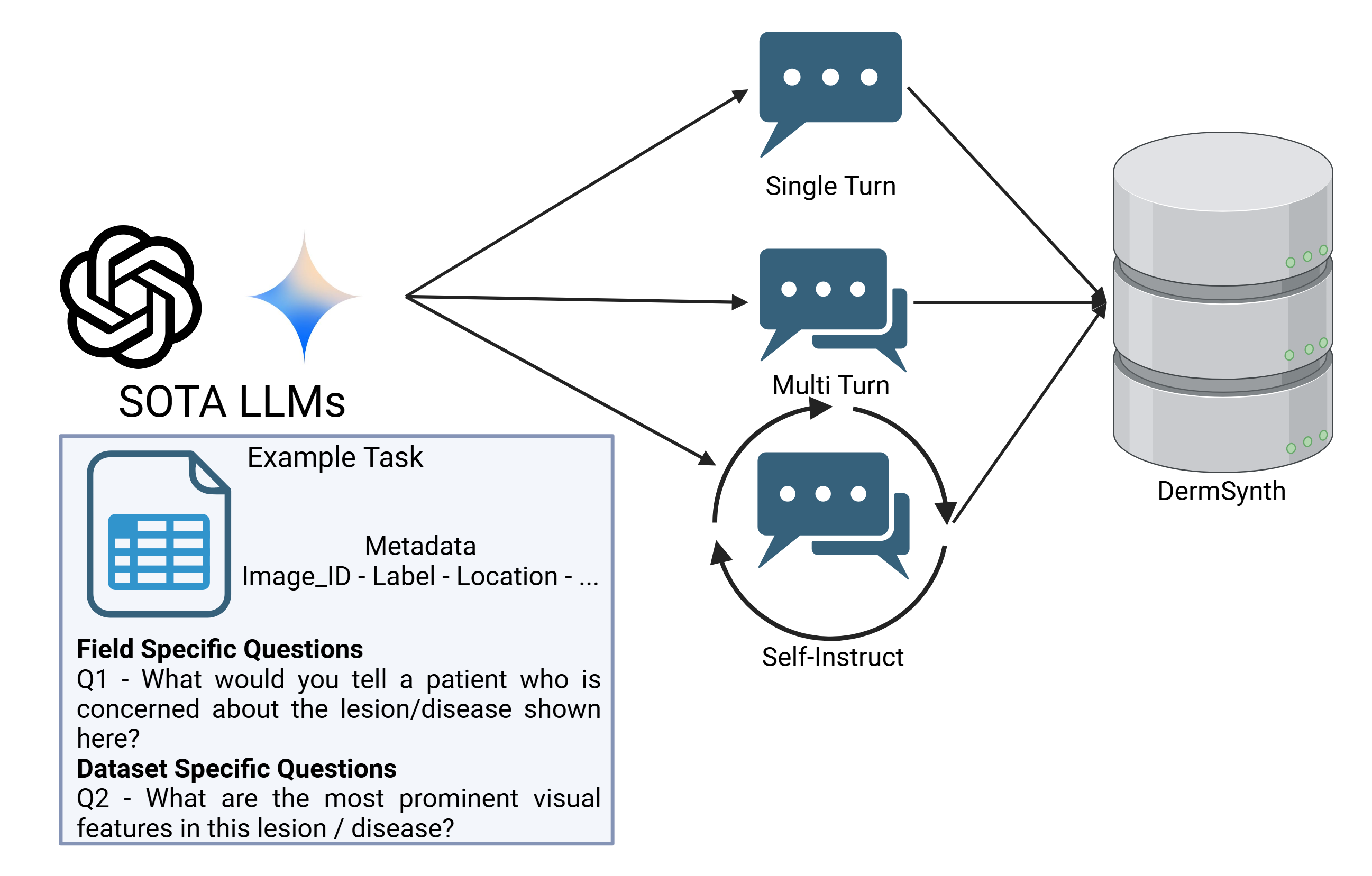}
    \caption{Overview of the synthetic data creation process for \emph{DermaSynth}. State of the art large language model (Gemini 2.0) were used to generate synthetic and clinically relevant image-text pairs\cite{fig1}.}
    \label{fig:graphical_abstract}
\end{figure}

\section*{2. Data}
We curated \emph{DermaSynth} using multiple open-access dermatological datasets. We used datasets with permissive CC-BY-4.0 licenses, thereby excluding any images with restrictive terms. We generated general and dataset specific questions using ChatGPT o1 - a state-of-the-art (SOTA) model (accessed on: 20 December 2024). We then combined general prompts and metadata-based prompts to synthesise realistic question--answer pairs, refining outputs thorough post-processing stage. The final dataset statistics, shown in Table~\ref{tab:dataset_stats}, underscore the breadth and diversity of this resource, which we intend to facilitate a range of dermatology-focused AI applications.\\

\noindent \textbf{Dataset Sources.}
We collected open-access dermatological images from repositories as follows:
\begin{itemize}
    \item DERM12345\cite{yilmaz2024derm12345} (CC-BY-4.0 License)
    \item BCN20000\cite{hernandez2024bcn20000} (CC-BY-4.0 License)
    \item PAD-UFES-20\cite{pacheco2020pad} (CC-BY-4.0 License)
    \item SCIN\cite{ward2024crowdsourcing} (CC-BY-4.0 License)
    \item HIBA\cite{ricci2023dataset} (CC-BY-4.0 License)
\end{itemize}
Datasets or images with more restrictive licenses (e.g., CC-BY-NC) were \textbf{excluded} to ensure that our final release aligns with open-access licenses. 

\begin{figure}[!ht]
    \centering
    \includegraphics[width=1\textwidth]{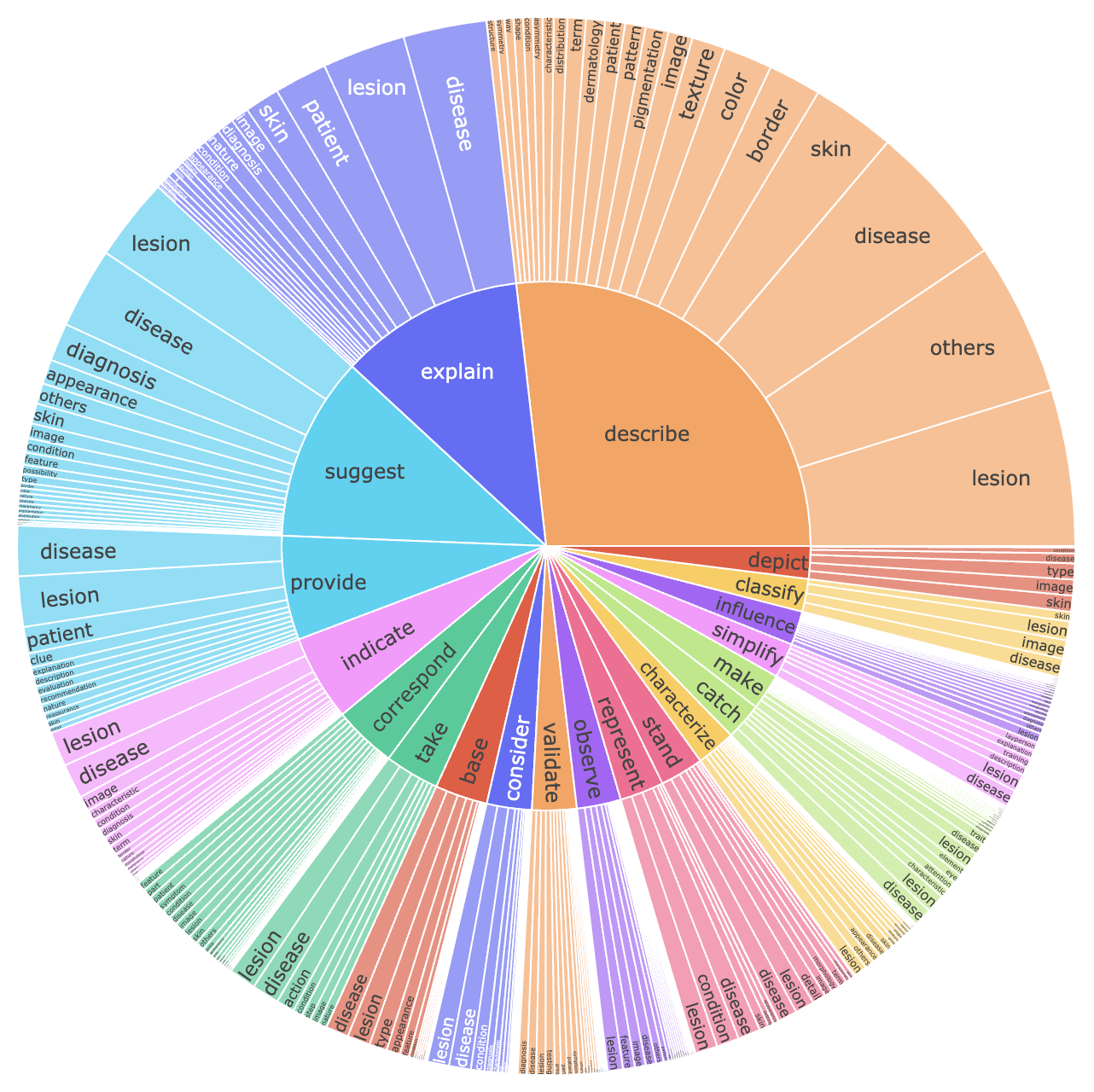}
    \caption{The most common 20 root verb-noun pairs of question.}
    \label{fig:root_verb}
\end{figure}

\vspace{1em}
\noindent \textbf{Question Prompt Generation.}
To ensure diverse, realistic prompts, we used a guiding instruction (Appendix 1 - Question Generation Prompt) that emphasised generating queries centered on the image rather than strictly referencing labels. This question generation prompt was used to generate question set using the ChatGPT o1 model. The most common 20 root verb-noun pairs of question set is shown in Figure \ref{fig:root_verb} which indicates the question set is diverse. Prompts spanned a wide range of question types—such as ``What does this lesion look like?'' and ``Could this be \textit{X}?''—and were manually screened and selected for clarity and medical relevance. This manual selection of prompt examples provides a dermatologist's investigative approach, allowing the synthetic Q\&A to cover nuanced scenarios without over-reliance on labels or assumptions.

\noindent \textbf{Synthetic Data Generation.}
We used two questions per image to generate synthetic image-text pairs as follows:

\begin{enumerate}
    \item \emph{General Prompt Question}: Drawn from a pool of question types (total 120 questions and 20 variations per type): Observational Overview, Ask for Diagnosis, Differential Diagnosis, Patient-Centered Explanation, Next Steps, and Creative Scenarios.
    
    \item \emph{Dataset Specific Metadata-Based Question}: Incorporates dataset metadata (e.g., label, patient age, anatomic site, skin type). 
\end{enumerate}

\begin{figure}[ht]
    \centering
    \includegraphics[width=1\textwidth]{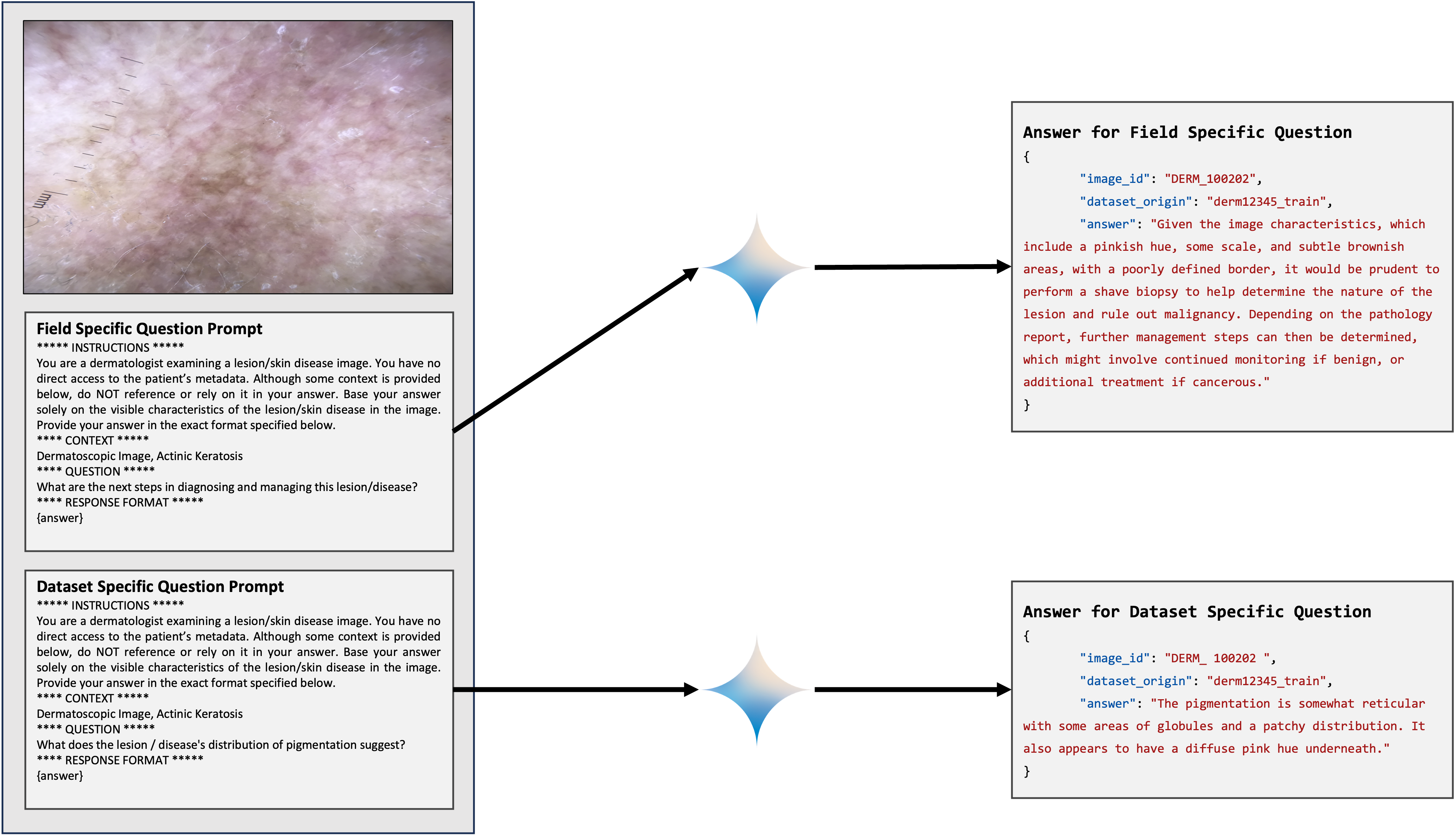}
    \caption{A figure from DERM12345 dataset with a field specific question and a dataset specific question with their Gemini 2.0 answers.}
    \label{fig:example}
\end{figure}

\noindent We generated DermaSynth by sending images with their questions and specialised prompt (Appendix 1 - API Prompt Instruction) to Gemini 2.0 API (model: gemini-2.0-flash-exp, accessed between: 10 January 2025 - 25 January 2025). After synthetic text generation, we post-process the data to remove unsuitable or nonsensical outputs such as hallucinated outputs. This ensures that the final dataset is both clinically relevant and consistent in style. In addition, some images from the SCIN dataset \cite{ward2024crowdsourcing} has additional symptom metadata (e.g., itching or scaling). To explore how symptom context affects model performance, we generated two variants of image--text pairs from these samples: one that explicitly includes symptom details and one that omits them. An example figure with questions and answers is shown in Figure \ref{fig:example}.

\noindent
\textbf{Self-Instruct Data Generation.}
A self-instruct approach was applied to create a wide variety of dermatology-focused tasks, starting with more than 50 seed prompts tailored to the images in each dataset. This approach helps to align AI models with target fields. These seed prompts were iteratively expanded through automated querying of large language models, producing both new questions and refined responses. This iterative process ensures that the generated instruction–response pairs capture a broad range of clinical perspectives, disease types and lesion types, while limiting overly repetitive or clinically irrelevant outputs.

\vspace{1em}
\noindent \textbf{Dataset Statistics.}
Table~\ref{tab:dataset_stats} shows an overview of each dataset's size and their corresponding image-text pair size. 

\begin{table}[ht]
\centering
\caption{Specifications of the datasets and their generated image-text pairs in \emph{DermaSynth}.}
\label{tab:dataset_stats}
\begin{tabular}{lrrrr}
\hline
\textbf{Dataset} & \textbf{Size} & \textbf{Clinical} & \textbf{Dermatoscopic} & \textbf{Image-Text Pairs} \\
\hline
DERM12345 - Train & 9,860 & - & 9,860 & 19,662 \\
DERM12345 - Test & 2,485 & - & 2,485 & 4,966 \\
BCN20000 - Train & 12,413 & - & 12,413 & 24,799 \\
BCN20000 - Test & 6,533 & - & 6,533 & 12,994 \\
PAD-UFES-20 & 2,298 & 2,298 & - & 4,591 \\
SCIN & 10,000 & 10,000 & - & 21,859 \\
HIBA & 1,616 & 346 & 1,270 & 3,149 \\
\hline
Total & 45,205 & 13,568 (30\%) & 32,561 (70\%) & 92,020 \\
\hline
\end{tabular}
\end{table}

\section*{3. Baseline Model}
\begin{figure}[!ht]
    \centering
    \includegraphics[width=1\textwidth]{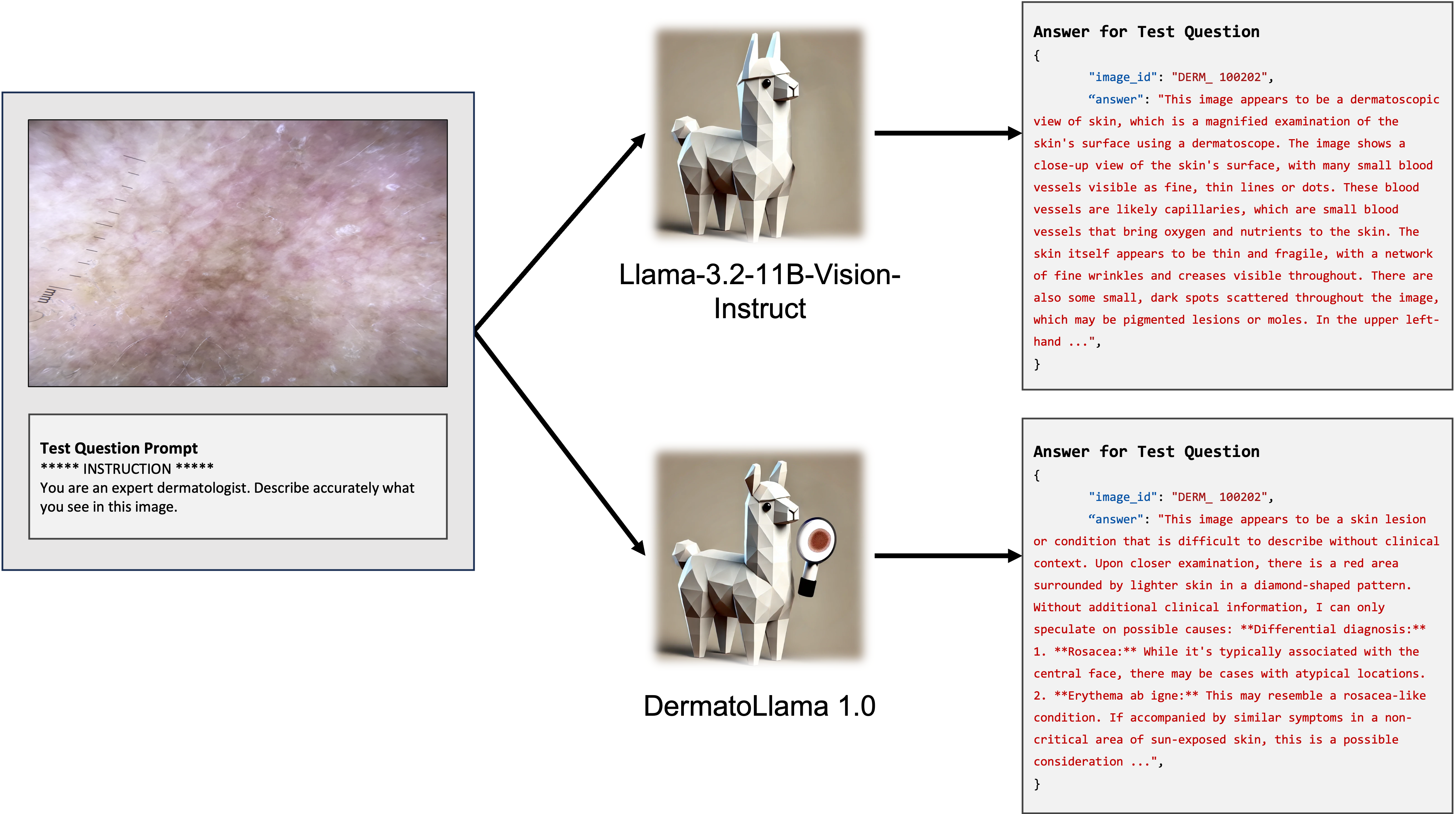}
    \caption{Shows input images with a simple prompt and their answers by original Llama model and DermatoLlama model. Both the standard llama illustration and the variant featuring a dermatoscope were generated using DALL·E 3.}
    \label{fig:test}
\end{figure}

\paragraph{Setup.}
For a proof-of-concept, we sampled 5,000 image--text pairs from the DERM12345 \cite{yilmaz2024derm12345} dataset. We then fine-tuned a preliminary \textbf{Llama-3.2-11B-Vision-Instruct} model (float16), DermatoLlama, using Python (v3.11.11) and the ``unsloth'' library (v2025.1.7) on a single NVIDIA A100 (40GB) instance with 96GB system RAM. This model is accessible from HuggingFace at \url{https://huggingface.co/abdurrahimyilmaz/DermatoLlama-1.0}. An example output of the original Llama model and the DermatoLlama model for a simple input is shown in Figure~\ref{fig:test}.

\section*{4. License and Non-Commercial Use}
\textbf{DermaSynth} and \textbf{DermatoLllama 1.0} are released strictly for academic and non-commercial use (CC-BY-NC 4.0). This restriction aligns with:
\begin{enumerate}
    \item \emph{Llama’s Non-Commercial License:} Since our model fine-tuning pipeline relies on Llama, any derivative work inherits its non-commercial policy.
    \item \emph{Upstream Model Constraints:} Some instruction data is based on OpenAI, Google's Gemini and other services with terms prohibiting competitive commercial usage.
    \item \emph{Safety and Maturity Considerations:} We have not instituted sufficient safety measures for broad clinical deployment. As a result, \textbf{DermaSynth} and \textbf{DermatoLllama 1.0} is not ready for general commercial use.
\end{enumerate}

\section*{5. Resources Released}
We release the following resources:
\begin{itemize}
    \item \textbf{DermatoLlama 1.0:} A fine-tuned Llama-3.2-11B-Vision-Instruct model trained on 5k samples from the DERM12345-train dataset \cite{yilmaz2024derm12345} with specialised dermatology knowledge. 
    \item \textbf{DermaSynth:} 92,020 synthetic image--text pairs with quality-checked in \texttt{Hugging Face Datasets} and JSON format.
    \item \textbf{Training Scripts:} Python utilities and a \texttt{Google Colab} notebook showcasing an end-to-end fine-tuning workflow with the \texttt{unsloth} library.
\end{itemize}

\noindent All resources can be accessed via \url{https://github.com/abdurrahimyilmaz/DermaSynth}. The model is accessible at \url{https://huggingface.co/abdurrahimyilmaz/DermatoLlama-1.0}. This dataset, model and codes are non-commercial use only, abiding by the policies noted above.

\section*{6. Discussion}
\noindent
The integration of LLMs in clinical workflows has a growing interest with its more responsible and explainable nature, especially within the field of dermatology \cite{zarfati2024exploring,cao2024mpoxvlm}. Here, models must not only achieve high success but also clearly present their reasoning to build trust among clinicians and patients. Factors such as bias monitoring, fairness checks, and interpretable decision paths have become critical for ensuring that AI-driven systems comply with ethical and clinical standards. As researchers increasingly explore using vision LLMs for decision support systems, placing emphasis on developing responsible AI practices will accelerate clinician engagement.

This study has some limitations. While synthetic data can help to train and fine-tune LLMs, such features are rarely represented in SOTA LLMs' knowledge bases/cutoffs. SOTA LLMs are mostly not trained on the medical corpora and latest developments in dermatology (Gemini 2.0 knowledge cutoff: August 2024 as of February 2025). Their training datasets and knowledge cutoffs might result in incomplete or not up-to-date synthetic data.  Models trained on synthetic data may therefore struggle to capture and explain subtle morphological nuances or rare lesion types/cases. Expert review of synthetic content remains essential but is resource-intensive (this work is currently underway). This drawback underscores the continuing need for expert-driven curation and validation. Moreover, relying exclusively on synthetic datasets risks overlooking real-world variability in lesion presentation, lighting conditions, and patient demographics.

Beyond these limitations, several avenues can strengthen both the framework and its practical relevance in the future. One direction is to design multi-agent systems, each specialising in a subfield of dermatology—such as rare diseases or pigmentary disorders, to enhance accuracy through collaborative validation. Another major consideration involves extended, multi-turn dialogues that integrate clinician feedback (e.g., via Reinforcement Learning from Human Feedback (RLHF)\cite{tanno2024collaboration} or Retrieval Augmented Generation (RAG)) to mimic diagnostic inquiries more accurately. In addition, training with larger and more diverse datasets can further refine performance on complex lesion types and underrepresented demographic groups, enabling models to adapt to the full spectrum of clinical scenarios.

Methods such as student–teacher model paradigms and knowledge distillation are efficient methods used in this work to facilitate more resource-efficient deployments. Broader quantitative and qualitative evaluations for these deployments are needed to determine how effectively these innovations align with evolving medical standards. Collaborations with dermatology experts will be critical: such partnerships not only offer in-depth insight into clinical outcomes, but also help to iteratively refine both synthetic datasets and the models tasked with interpreting them. 

By transferring expertise from large instruction-tuned language-and-vision models to smaller, specialised networks, we released a rich dataset (DermaSynth) and a scalable model (DermatoLlama 1.0) that can be used with low-cost and open-source systems, enabling researchers to accelerate LLM research in the dermatology field.

\section*{Acknowledgments}
Abdurrahim Yilmaz has been funded by the President's PhD Scholarships at Imperial College London. Donghee Choi and Joram M. Posma are supported by the Horizon Europe project CoDiet. The CoDiet project is funded by the European Union under Horizon Europe grant number \href{https://doi.org/10.3030/101084642}{101084642}. CoDiet research activities taking place at Imperial College London is supported by UK Research and Innovation (UKRI) under the UK government's Horizon Europe funding guarantee (grant number \href{https://doi.org/10.3030/101084642}{101084642}).

\bibliographystyle{unsrt}
\bibliography{bibliography}

\begin{thebibliography}{10}

\bibitem{peng2023instruction}
Baolin Peng, Chunyuan Li, Pengcheng He, Michel Galley, and Jianfeng Gao.
\newblock Instruction tuning with gpt-4.
\newblock {\em arXiv preprint arXiv:2304.03277}, 2023.

\bibitem{liu2024visual}
Haotian Liu, Chunyuan Li, Qingyang Wu, and Yong~Jae Lee.
\newblock Visual instruction tuning.
\newblock {\em Advances in neural information processing systems}, 36, 2024.

\bibitem{liu2024improved}
Haotian Liu, Chunyuan Li, Yuheng Li, and Yong~Jae Lee.
\newblock Improved baselines with visual instruction tuning.
\newblock In {\em Proceedings of the IEEE/CVF Conference on Computer Vision and Pattern Recognition}, pages 26296--26306, 2024.

\bibitem{wu2024pmc}
Chaoyi Wu, Weixiong Lin, Xiaoman Zhang, Ya~Zhang, Weidi Xie, and Yanfeng Wang.
\newblock Pmc-llama: toward building open-source language models for medicine.
\newblock {\em Journal of the American Medical Informatics Association}, page ocae045, 2024.

\bibitem{han2023medalpaca}
Tianyu Han, Lisa~C Adams, Jens-Michalis Papaioannou, Paul Grundmann, Tom Oberhauser, Alexander L{\"o}ser, Daniel Truhn, and Keno~K Bressem.
\newblock Medalpaca--an open-source collection of medical conversational ai models and training data.
\newblock {\em arXiv preprint arXiv:2304.08247}, 2023.

\bibitem{shu2023visual}
Chang Shu, Baian Chen, Fangyu Liu, Zihao Fu, Ehsan Shareghi, and Nigel Collier.
\newblock Visual med-alpaca: A parameter-efficient biomedical llm with visual capabilities, 2023.

\bibitem{li2024llava}
Chunyuan Li, Cliff Wong, Sheng Zhang, Naoto Usuyama, Haotian Liu, Jianwei Yang, Tristan Naumann, Hoifung Poon, and Jianfeng Gao.
\newblock Llava-med: Training a large language-and-vision assistant for biomedicine in one day.
\newblock {\em Advances in Neural Information Processing Systems}, 36, 2024.

\bibitem{alpaca}
Rohan Taori, Ishaan Gulrajani, Tianyi Zhang, Yann Dubois, Xuechen Li, Carlos Guestrin, Percy Liang, and Tatsunori~B. Hashimoto.
\newblock Stanford alpaca: An instruction-following llama model.
\newblock \url{https://github.com/tatsu-lab/stanford_alpaca}, 2023.

\bibitem{selfinstruct}
Yizhong Wang, Yeganeh Kordi, Swaroop Mishra, Alisa Liu, Noah~A. Smith, Daniel Khashabi, and Hannaneh Hajishirzi.
\newblock Self-instruct: Aligning language model with self generated instructions, 2022.

\bibitem{fig1}
Abdurrahim Yilmaz and Burak Temelkuran.
\newblock Created in biorender.
\newblock \url{https://BioRender.com/f99q320}, 2025.

\bibitem{yilmaz2024derm12345}
Abdurrahim Yilmaz, Sirin~Pekcan Yasar, Gulsum Gencoglan, and Burak Temelkuran.
\newblock Derm12345: A large, multisource dermatoscopic skin lesion dataset with 40 subclasses.
\newblock {\em Scientific Data}, 11(1):1302, 2024.

\bibitem{hernandez2024bcn20000}
Carlos Hern{\'a}ndez-P{\'e}rez, Marc Combalia, Sebastian Podlipnik, Noel~CF Codella, Veronica Rotemberg, Allan~C Halpern, Ofer Reiter, Cristina Carrera, Alicia Barreiro, Brian Helba, et~al.
\newblock Bcn20000: Dermoscopic lesions in the wild.
\newblock {\em Scientific Data}, 11(1):641, 2024.

\bibitem{pacheco2020pad}
Andre~GC Pacheco, Gustavo~R Lima, Amanda~S Salomao, Breno Krohling, Igor~P Biral, Gabriel~G de~Angelo, F{\'a}bio~CR Alves~Jr, Jos{\'e}~GM Esgario, Alana~C Simora, Pedro~BC Castro, et~al.
\newblock Pad-ufes-20: A skin lesion dataset composed of patient data and clinical images collected from smartphones.
\newblock {\em Data in brief}, 32:106221, 2020.

\bibitem{ward2024crowdsourcing}
Abbi Ward, Jimmy Li, Julie Wang, Sriram Lakshminarasimhan, Ashley Carrick, Bilson Campana, Jay Hartford, Tiya Tiyasirichokchai, Sunny Virmani, Renee Wong, et~al.
\newblock Crowdsourcing dermatology images with google search ads: Creating a real-world skin condition dataset.
\newblock {\em arXiv preprint arXiv:2402.18545}, 2024.

\bibitem{ricci2023dataset}
Mar{\'\i}a~Agustina Ricci~Lara, Mar{\'\i}a~Victoria Rodr{\'\i}guez~Kowalczuk, Maite Lisa~Eliceche, Mar{\'\i}a~Guillermina Ferraresso, Daniel~Roberto Luna, Sonia~Elizabeth Benitez, and Luis~Daniel Mazzuoccolo.
\newblock A dataset of skin lesion images collected in argentina for the evaluation of ai tools in this population.
\newblock {\em Scientific Data}, 10(1):712, 2023.

\bibitem{zarfati2024exploring}
Mor Zarfati, Girish~N Nadkarni, Benjamin~S Glicksberg, Moti Harats, Shoshana Greenberger, Eyal Klang, and Shelly Soffer.
\newblock Exploring the role of large language models in melanoma: A systematic review.
\newblock {\em Journal of Clinical Medicine}, 13(23):7480, 2024.

\bibitem{cao2024mpoxvlm}
Xu~Cao, Wenqian Ye, Kenny Moise, and Megan Coffee.
\newblock Mpoxvlm: A vision-language model for diagnosing skin lesions from mpox virus infection.
\newblock {\em arXiv preprint arXiv:2411.10888}, 2024.

\bibitem{tanno2024collaboration}
Ryutaro Tanno, David~GT Barrett, Andrew Sellergren, Sumedh Ghaisas, Sumanth Dathathri, Abigail See, Johannes Welbl, Charles Lau, Tao Tu, Shekoofeh Azizi, et~al.
\newblock Collaboration between clinicians and vision--language models in radiology report generation.
\newblock {\em Nature Medicine}, pages 1--10, 2024.

\end{thebibliography}

\newpage            
\appendix          
\section*{Appendix 1: Prompts}

\vspace{1em}
\noindent \textbf{Question Generation Prompt}
For question generation prompt, we used that prompt by giving metadata information to ChatGPT o1:

\begin{myshadedbox}
\noindent
***** INSTRUCTIONS *****\\
Write questions to generate synthetic data using large language models. For generalizability, we need to create more prompt samples so while using the API to generate the synthetic dataset, we can randomly sample from any prompt example. Lets continue enhancing the ``XXXX'' dataset prompts. Generate more prompt types, start from 1. (you can use the current ones if you'd like). 
Only thing available to us are images, labels and metadata. We can not assume somethings about the images and we will not annotate anything. 
The prompt should be centered about the image, not the label. 
Note that we do not always have to give the label and ask a question, we can create prompt types that asks for the label, metadata etc. depending on the dataset. This simulates a dermatologist using the chatbot inferencing an image by asking ``What is this image'', ``What type of... is this image?'' like questions. 
\\\\
***** DATASET CONTEXT *****\\
\{metadata\_context\} (e.g. Anatomical site - Sex - Skin Type)

\end{myshadedbox}

\vspace{1em}
\noindent \textbf{API Prompt Instruction.}
During synthetic data generation, we use the following instruction prompt to focus on visible lesion characteristics and not reveal or rely on metadata:

\begin{myshadedbox}
\noindent
***** INSTRUCTIONS *****\\
You are a dermatologist examining a lesion/skin disease image.\\
You have no direct access to the patient's metadata. Although some context is provided below, do NOT reference or rely on it in your answer. Base your answer solely on the visible characteristics of the lesion/skin disease in the image.\\
Provide your answer in the exact format specified below.
\\\\
***** CONTEXT *****\\
\{metadata\_context\} (e.g. Label: Actinic keratosis - Age: 70.0 - Anatomical Site: head/neck -Sex: female)
\\\\
***** QUESTION *****\\
\{prompt\_text\}
\\\\
***** RESPONSE FORMAT *****\\
\{answer\}
\end{myshadedbox}

\newpage            
\section*{Appendix 2: Example Questions}
All questions are accessible at code repositories.

\vspace{1em}
\noindent \textbf{Example General Questions}

\begin{myshadedbox}
\begin{itemize}
    \item What are the most prominent visual features in this lesion/disease?
    \item Based on the image, what might be the diagnosis?
    \item What conditions might present similarly to the lesion/disease in this image?
    \item How would you rank the most likely alternative diagnoses for this lesion/disease?
    \item How would you explain the lesion/disease in this image to a worried patient?
    \item What language would you use to describe this lesion/disease in a patient consultation?
    \item What follow-up procedures might be necessary for this lesion/disease?
    \item What further evaluations are essential for this lesion/disease's treatment or diagnosis?
    \item Develop a question--answer set for teaching purposes about this lesion/disease.
    \item Write a concise Q\&A pair for explaining the lesion/disease in this image.
\end{itemize}
\end{myshadedbox}

\noindent \textbf{Example Dataset Specific Questions (for the DERM12345 dataset \cite{yilmaz2024derm12345})}

\begin{myshadedbox}
\begin{itemize}
    \item Based on the appearance of this lesion/skin disease, what condition might it represent?
    \item Considering its morphology, what might this lesion/disease indicate?
    \item What does the lesion / disease's distribution of pigmentation suggest?
    \item Could this lesion/skin disease represent a malignant condition? Why or why not?
    \item What dermatological disorders might resemble the lesion/skin disease in this image?
    
\end{itemize}
\end{myshadedbox}

\end{document}